\documentclass[11pt]{article}

\usepackage[preprint]{acl}

\usepackage{times}
\usepackage{latexsym}

\usepackage[T1]{fontenc}

\usepackage[utf8]{inputenc}

\usepackage{microtype}

\usepackage{inconsolata}

\usepackage{graphicx}

\usepackage{amsmath}
\usepackage{booktabs}
\usepackage{comment}
\usepackage{makecell}

\usepackage{listings}
\usepackage{adjustbox}

\usepackage{fancyhdr}

\usepackage{hugo-defs}
\title{\benchmarkname: The SciFy Scientific Feasibility Benchmark
}

\author{
Cash Costello,
James Mayfield, 
Elsbeth Turcan, 
Christine Piatko, 
Christina K. Pikas, \\
\bf Justin Rokisky, 
Sam Scheck, 
Chris Ribaudo, 
Ritwik Bose, 
Alex Memory\\
Johns Hopkins Applied Physics Laboratory\\
11100 Johns Hopkins Rd.\\
Laurel, MD 21045
}

\usepackage[hang,flushmargin]{footmisc}

\begin{document}

\maketitle

\begingroup
\renewcommand{\thefootnote}{}
\footnotetext{DISTRIBUTION STATEMENT A. Approved for public release: distribution is unlimited. This material is based upon work supported by the DARPA Scientific Feasibility (SciFy) Program under Contract HR001122D0001. Any opinions, findings and conclusions or recommendations expressed in this material are those of the authors and do not necessarily reflect the views of the Department of War or the U.S. Government.}
\endgroup

\begin{abstract}

We present \benchmarkname, a benchmark dataset for evaluating systems that assess the feasibility of scientific claims.
\benchmarkname{} includes \numclaims{} claims in materials science,
each annotated with a ground-truth feasibility score on a five-point scale
along with an explanation of that assessment.
The collection differs from previous collections in several important ways:
1) it defines a complex task that requires reasoning over
claims of varying scientific feasibility;
2) its claims are not extracted from existing scientific publications
but are created de novo,
greatly reducing the chances that LLMs have trained on them;
3) claims and ground truth are established by subject matter experts, not by artificial intelligence; and
4) unlike many benchmarks that ask about question/answer pairs,
provide multiple choice answers,
or ask questions requiring short, fixed answers,
\benchmarkname{} explanations are completely open-ended.
We describe the benchmark design, data creation process, and evaluation metrics, and we report baseline results using recent GPT models.

\end{abstract}

\section{Introduction}

This paper describes \benchmarkname, a new benchmark for systems that assess feasibility of scientific claims. A scientific claim such as ``a perovskite solar cell of ABX3 composition was formed with a measured bandgap of 1.7 eV,''
is \textit{feasible} if it could be achieved, given appropriate resources, by a sufficiently skilled team using current scientific knowledge and technology.
A \textit{feasibility assessment} is an estimate of that feasibility, expressed as a score
together with a supporting explanation.
\benchmarkname{} comprises a set of scientific claims in the domain of materials science
together with ground truth feasibility assessments of those claims
that can be used to compare the effectiveness
of systems that are evaluated on these claims.

In practice, feasibility assessment is typically performed by human experts, who synthesize evidence from prior literature, theoretical constraints, and available technologies to judge whether a claim can be realized.
This process is time-consuming, requires deep domain expertise, and does not scale well as the volume of scientific claims increases.
These challenges motivate the development of automated systems that can assist with, or partially automate, scientific feasibility assessment.

As scientific claims proliferate in press releases, social media, and scientific publications, the ability to assess their feasibility is increasingly important.
Accurate feasibility assessment is important for research prioritization, investment evaluation, technology forecasting, and policy planning.
At the same time, infeasible claims can arise for many reasons. A claimant might mistakenly believe a claim to be true.
For example, a journalist may misunderstand the results of a scientific paper, or a scientist may overlook a confounding factor or calculation error.
Claims can also originate from large language models, which may hallucinate plausible-sounding but infeasible statements.
Regardless of the source, distinguishing feasible claims from infeasible ones is essential for sound scientific and technological decision-making.

\benchmarkname{} was developed under the DARPA SciFy program.
The objective of the DARPA Scientific Feasibility (SciFy) program
is ``to develop computational methods that measure the feasibility of claims
in order to enable accurate assessments of scientific content''~\cite{scifyBAA}
The program’s focus is on claims that express scientific and technological capabilities.
The program aims ``to demonstrate that the scientific feasibility of claims can be determined
by using automated reasoning to decompose claims into constituent, verifiable parts.''
To evaluate feasibility assessment systems developed under the program's funding, Technical Area 2 of DARPA SciFy is focused on developing original datasets that are sufficiently large, of high quality, and support multiple types of reasoning tasks.  \benchmarkname{} is the result of the first round of evaluation of the systems participating in the program.

\benchmarkname{} includes \numclaims{} claim problems with ground truth for each.
Ground truth has two components:
a feasibility score drawn from a -2 to 2 Likert scale
indicating a subject matter expert's (SMEs') assessment of the claim's feasibility,
and an explanation of why the SME came to that conclusion.
Feasibility scores can be automatically graded,
thereby allowing systems to be compared for effectiveness.
Automatic scoring of explanations 
remains an open problem
for two reasons.
First, there may be more than one acceptable explanation for a claim's feasibility,
only one of which will have been captured in the ground truth provided by the SME.
Second, explanations are complex semantic objects,
and while large language models have some ability to do semantic comparison,
they can struggle with such tasks,
especially for technical areas.

Our contributions include a novel benchmark dataset for AI reasoning, accompanying ground truth and scripts for calculating metrics, and results of using \benchmarkname{} to evaluate modern AI systems revealing that scientific claim assessment is a challenging task.

\section{Related Work}

\subsection{Evaluation of Reasoning Systems}

\benchmarkname{} is a member of what might be called the class of reasoning system evaluations.
This is a broad class that varies along several important dimensions.

\paragraph{Task Complexity.}
Independent of the reasoning that may be required,
systems vary in the complexity of the task presented to the systems being evaluated.
The simplest tasks to evaluate require true/false or multiple choice responses.
For example, Google-Proof Question Answering (GPQA) (Diamond subset)~\citep{GPQA}
is a multiple-choice question benchmark in biology, physics, and chemistry
requiring significant domain expertise.
It is designed such that web search is minimally helpful,
allowing it to probe internal reasoning over search and memorization.
Massive Multitask Language Understanding (MMLU-Pro)~\citep{mmlu-pro}
is a large multiple-choice question benchmark across science and humanities domains,
focusing on reasoning-intensive questions with many multiple-choice options;
it is an evolution of the original MMLU benchmark
for which results are reported for many LLMs.

Task complexity can vary along several dimensions,
including multilinguality, multimodality, and answer delivery.
MultiNRC~\cite{multinrc} assesses LLMs in Chinese, French, and Spanish.
Multilingual Grade School Math (MSGM)~\cite{mgsm} assesses reasoning on grade school mathematics in eleven languages.
MuBench~\cite{mubench} tests an LLM's abilities in 61 languages,
including a variety of lower resource languages that LLMs tend to struggle with.

Many scientific benchmarks are now multimodal, often targeting a broad range of disciplines.
Humanity's Last Exam (HLE)~\citep{HLE}
is a multimodal benchmark across a wide range of fields,
created in response to the saturation of older benchmarks
and designed specifically such that frontier models at the time of its publication
(e.g., OpenAI~o1) could not solve its questions.
The Abstraction and Reasoning Corpus for Artificial General Intelligence 2 (ARC-AGI-2)~\citep{arcagi2}
is a multimodal benchmark and successor to the original ARC that tests abstract reasoning capabilities;
it remains challenging for state-of-the-art models.

Answer delivery varies across reasoning evaluation datasets.
Three broad classes are A) you run the evaluation in your environment;
B) you run the evaluation in a docker container provided by the evaluators
to restrict the available resources; and
C) you submit your containerized system and the evaluator runs it on sequestered data.
\benchmarkname{} and most of the above test sets fall in Class A,
typically taking system output as JSON.
Some evaluations need to ensure that all evaluated systems operate in the same context.
Examples of this class of benchmark include
BigCodeBench~\cite{bigcodebench}, which assesses coding ability,
SWE-Bench~\cite{swebench}, which assesses reasoning about code,
GAIA~\cite{gaia}, which assesses multimodal reasoning,
and METR~\cite{metr}, which assesses ability to identify safety risks.
Sometimes it is important to sequester the test data.
In such cases, the system itself is submitted, usually as a Docker container.
Grand Challenge-based evaluations such as the RARE25 Challenge~\cite{rare25},
which evaluates reasoning about early-stage cancer,
the F1TENTH Competition~\cite{f1tenth}, which evaluates path planning,
and ARC-AGI variants~\cite{arcagi2}, which assess visual reasoning,
all operate in this manner.

\paragraph{Reasoning Complexity.}

Complex reasoning benchmarks that test a system's inference abilities are also available
(e.g., fact-checking~\cite{averitec} and general reasoning~\cite{arcagi2}),
albeit not specifically in scientific domains.
Scientific claim verification benchmarks that require deep reasoning
and do not assume solutions to their test cases are rarer.
FEVER~\cite{fever} for example creates claims by modifying Wikipedia sentences,
then independently determining whether the resulting claim is supported, refuted,
or undecidable due to lack of information.
The 2025 American Invitational Mathematics Examination (AIME 2025)~\citep{aime}
is a collection of real mathematics test questions given to human participants attempting to qualify for the USA Mathematical Olympiad, a selective high school mathematics competition.

Others are devoted to more technical domains.
PhyX~\citep{PhyX} is a multimodal benchmark based specifically in physics understanding,
requiring domain expertise and multi-step reasoning.
Massive Multi-Modal Reasoning (MMLU-Reason/MMMR)~\citep{MMMR}
is a reasoning benchmark for scientific, mathematical, and planning tasks,
with evaluations that analyze the type of logic errors made by the model.
Reasoning Bench (R-Bench)~\citep{R-Bench} is a graduate-level benchmark
combining text and multimodal questions across eighteen scientific disciplines.
This benchmark is explicitly multilingual, in English and Chinese.

While finding a solution to a scientific problem might tell you something about problem feasibility,
it does not test a system's ability to recognize infeasibility,
much less to argue the case for that infeasibility.

\paragraph{Verification.}

Benchmarks in which a statement or other artifact is presented,
asking the system to assess the veracity of the item,
are called \textit{verification tasks.}
Dmonte et al.\ present a list of more than 25 English language fact verification evaluation datasets developed from 2017 to 2024.
\footnote{\href{https://github.com/LanguageTechnologyLab/Claim-Verification-Papers}{github.com/LanguageTechnologyLab/Claim-Verification-Papers}}
Many such datasets are aimed at detection of misinformation,
and are built on extracted claims.
\citet{misinformation-guide} surveys 75 misinformation datasets.

Foundational work in scientific NLP established the task of verifying claims against existing literature.
The SciFact benchmark~\cite{scifact_2020} introduced a framework for identifying supporting or refuting evidence
within expert-written biomedical abstracts.
Subsequent iterations, such as SciFact-Open~\cite{scifact_open_2022} and MultiVerS~\cite{wadden2022multivers},
scaled this to open-domain retrieval and document-level reasoning.
COVID-Fact~\cite{covidfact} is an example of a similar dataset that is more focused on a particular scientific domain.
However, these systems are primarily retrospective,
evaluating claims whose truth is already established in the available corpus.
In contrast, the Covid VERification dataset (CoVERt)~\cite{covert} is a benchmark
consisting of fifteen medical claims,
96 abstracts with ``support'' or ``refute'' labels,
and a set of sentences extracted from the abstracts containing rationales for claim verification.

In the work closest to \benchmarkname{},
\citet{jansen-etal-2025-matter} developed a claim verification dataset called Matter-of-Fact
comprising 8.4K claims extracted from scientific articles.
In addition to claims attested in the articles,
Matter-of-Fact includes infeasible claims derived by modifying attested claims.
All judgments are binary:
a claim is feasible or it is infeasible.
This contrasts with \benchmarkname{},
which assigns a feasibility score on a $-2$ to $+2$ Likert scale.
All Matter-of-fact judgments are made by LLMs,
while \benchmarkname{} judgments are made by subject matter experts.
100 of Matter-of-Fact's claim scores were validated by a domain generalist,
but not by experts in the problem subdomains.
Thus, Matter-of-Fact is an excellent resource to help train a claim feasibility
or claim feasibility assessment tool,
but provides less nuanced and less certain feasibility judgments than \benchmarkname{}.

\paragraph{Solvability.}

Most reasoning system benchmarks assume the given task is solvable
and test the system's ability to find that solution.
As an example, FrontierScience~\cite{frontierscience} consists of PhD-level open-ended problems manually-created by human experts.
FrontierScience covers a broad range of scientific topics.
The task requires significant reasoning to achieve good performance.
However, the problems are still expressed as questions with known answers.
\benchmarkname{}'s claims in contrast comprise both claims that can be shown to be feasible
(i.e., have a solution)
and those that are infeasible
(i.e., have no known solution).

It is increasingly important for LLMs to be able to respond
that they cannot answer a given question.
AbstentionBench~\cite{abstentionbench} is a benchmark that includes 35K unanswerable questions.
UnsolvableQA is a framework that includes questions marked as unsolvable because they contain inherent contradictions, missing context, or ill-posed premises.
SQuAD 2.0~\cite{squad20} augments the original SQuAD benchmark with over 50,000 unanswerable questions written adversarially by crowdworkers to look like answerable ones.
A major advantage of this type of benchmark
is that it exposes an LLM's propensity to hallucinate answers.

\subsection{Approaches to Task Creation}

Test example creation is one of the most problematic aspects
of the development of reasoning system evaluations.
Test examples can be found or generated.
Finding test examples involves mining a selected resource
to identify in it instances of the target of the evaluation.

Example generation requires the generator to have native reasoning capability.
From an accuracy perspective,
people (at least for now) are the best choice for reasoner.
From an efficiency perspective though,
machine-generated examples are to be preferred.
People are expensive.
Most human-generated test sets contain only enough test examples
to detect effectiveness differences between systems to a desired level of significance.
A few benchmarks, such as MS~MARCO~\cite{msmarco} and Humanity's Last Exam~\cite{HLE}
contain large numbers of examples,
but at the expense of thousands or tens of thousands of human labor hours.
In contrast, machine-generated examples can typically be created at scale.
The downside of machine generation is that
establishing the correctness or validity of the examples
suffers from the deficiencies of the generating system.
Benchmarks based on LLM parametric memory generate flawed test examples due to
hallucination,
incorrect training data,
and incomplete training data.
Benchmarks try to handle these problems
by including an outside source of information within the example generation prompt.

Many benchmarks mine Wikipedia for such instances.
SQuAD~\cite{squad} manually mined 100,000 question/answer pairs from Wikipedia.
HotpotQA~\cite{hotpotqa} created questions whose answer could only be found
by examining multiple Wikipedia pages.
A major problem with using Wikipedia as a source of test examples
is that most LLMs will almost certainly contain all of Wikipedia in their training data.
This can lead to regurgitation from parametric memory
instead of demonstration of reasoning ability. 
Other benchmarks mine resources more closely aligned with the task they are trying to test.
For example, SWE-bench~\cite{swebench} mines GitHub issues and pull requests.
LegalBench~\cite{legalbench} mines US state and federal statutes and judicial opinions.
CURIE~\cite{curie} mines ArXiv and Peer-reviewed journals in Condensed Matter Physics and Materials Science.
MedQA~\cite{medqa} mines professional medical board exams.
FinQA~\cite{finqa} mines earnings reports.
The inclusion of these resources in LLM training data is less clear,
but likely still a problem.

One way to address the LLM training data problem
is to attach a date to each benchmark or each benchmark problem
and require that systems attempting the problem use only information prior to the given date.
For example, \citet{jansen-etal-2025-matter} attach a ``knowledge cutoff date''
to each component of the benchmark. 
This approach places the burden on the user of the benchmark
to ensure that the system being tested is reasoning
and not simply relying on parametric memory.

The alternative to finding test examples is to generate them synthetically.
For example, KOR-Bench~\cite{korbench} generates five types of reasoning problems
that rely on newly-defined rules and frameworks.
CLUTRR~\cite{clutrr} generates stories that contain family relationships to be discovered by the reasoning system.
SynLogic~\cite{synlogic} generates novel reasoning problems that include 35 varied reasoning tasks.
DivLogicEval~\cite{divlogiceval} maps symbolic logic propositions with potential answer sets
onto question types that are then converted to natural language problems.
MDBench~\cite{mdbench} generates synthetic document sets to assess a system's ability to reason across documents.

In practice, the distinction between found and generated test examples is blurred,
as at least some reasoning is always required
to identify promising candidates in the resource being mined
and marshal those candidates into benchmark examples.

\subsection{Other Related Work}

Automated scientific discovery has become a vibrant area of research
since the advent of large generative language models~\cite{baek-etal-2025-researchagent,
ghafarollahi2024sciagentsautomatingscientificdiscovery,
gottweis2025aicoscientist,
lehr2024chatgptresearchscientistprobing,
li2024chainideasrevolutionizingresearch,
lu2024aiscientistfullyautomated,
schmidgall2025agentlaboratoryusingllm,
si2024llmsgeneratenovelresearch,
su-etal-2025-many,
ueda2025exploringdesignmultiagentllm,
wang-etal-2024-scimon,
yamada2025aiscientistv2workshoplevelautomated}).
While the reasoning that goes on in these systems is much like reasoning for feasibility assessment,
these are not benchmarks
and their task is generative, not analytic.
\citet{SDE} provide a scientific discovery benchmark
that explores the ability of an LLM to contribute to scientific discovery.

Mined annotations of scientific papers provide a valuable resource both for benchmark creation
and for feasibility assessment system development.
While it is not itself a benchmark,
the Withdrar{X}iv~\citep{rao2024withdrarxiv} collection
contains over 14,000 scientific papers withdrawn from ar{X}iv
labeled with the reason for the retraction.
C{\small{LAIM}}C{\small{HECK}}~\cite{claimcheck_2025} is a dataset of NeurIPS 2023 and 2024 submissions
tied to weaknesses of those papers mined from OpenReview.
It includes ground truth for three annotation tasks:
associating weaknesses with claims,
classifying weaknesses relative to a fine-grained taxonomy,
and verifying a paper's claims with grounded reasoning.

\subsection{What Sets \benchmarkname{} Apart}
In summary, \benchmarkname{} includes a low complexity main task (feasibility scoring) and a high complexity explanation task. The reasoning complexity required to successfully complete the benchmark is high.  Feasibility scoring is a prospective verification task, while the explanation component is generative. \benchmarkname{} includes unsolvable claims. Test cases are created by people, who generate new examples that are not likely to be included in LLM training data. These features allow \benchmarkname{} to provide a challenging task that is likely to remain a solid evaluation vehicle for some time.

\section{Benchmark}

\subsection{Desiderata}

A benchmark for evaluating feasibility assessment systems must meet the following criteria:
\begin{itemize}
\itemsep0pt
    \item \textbf{Claims}:
    The claims that will be assessed for feasibility must be challenging enough
    to distinguish systems that rely solely on parametric memory and simple reasoning
    from those that can reason over unsound or incomplete data.

    \item \textbf{Feasibility Assessments}:
    A feasibility assessment estimates the likelihood that,
    given appropriate resources,
    the claim can be achieved
    using current scientific knowledge and current technology.  A feasibility assessment is expressed as a score
together with a supporting explanation.
    Because LLMs used as judges are known to exhibit ``self-preference bias''
    in which they favor systems built around their family of LLMs~\cite{judging-llm-as-judge,self-preference-bias},
    it is best that feasibility is assessed by human subject matter experts.
    It is also important for the use of metrics such as weighted Cohen's kappa
    (see Section~\ref{sec:metrics})
    that the distribution across feasibility values is not highly skewed.

    \item \textbf{Explanations}:
    Simply providing feasibility assessments, while useful and sufficient to produce a benchmark score,
    gives little insight into how those assessments were determined
    and where they might be right or wrong.
    A benchmark should therefore include the reasoning behind each feasibility assessment.
    It is not possible to provide a comprehensive set of explanations
    because there may be many reasoning chains that lead to a particular conclusion about feasibility.
    Nonetheless it is important to include the assessor's initial reasoning
    both to allow other experts to verify the conclusions
    and to provide a basis for semi-automated explanation evaluation.

    \item \textbf{Metrics}:
    It is important that results of the benchmark over time are comparable.
    To that end, there should be standard implementations of the metrics
    users of the benchmark are encouraged to use.

    \item \textbf{Time Stamp}:
    Claim feasibility can change over time.
    A claim that is infeasible at one time
    can become feasible with the introduction of a new technology or insight.
    Thus, a feasibility benchmark must specify the date on which each claim was assessed.
    Proper use of the benchmark requires the system being evaluated
    to restrict the information on which it bases its assessments
    to the information available at the specified time.
    Benchmarks that rely on extracted claims for their ground truth instead of human judgments
    are forced to use the date the claim was made as the assessment date.
    If human experts create the ground truth,
    the time stamp for everything in the collection can be the time of ground truth creation.
    
    \item \textbf{Baselines}:
    It is helpful to future researchers to provide baseline results on the benchmark data,
    either as a pointer to where the results might be found
    or, preferably, with the benchmark distribution itself.

\end{itemize}

\subsection{Benchmark Contents}

The \benchmarkname{} dataset consists of \numclaims{} scientific claims written by SMEs in materials science.
The task is to predict the feasibility of each claim and provide explanations for those feasibility assessments.
An example claim and explanation looks like this:
\begin{quote}\small
\textbf{Claim:} V--Fe--Co--Ni--Pt high-entropy alloys with an A15 phase have been synthesized,
exhibiting superconductivity with critical temperatures up to 5.18\,K.

\textbf{Feasibility:} $-2$

\textbf{Explanation:}
\begin{itemize}\setlength{\topsep}{0pt}\setlength{\itemsep}{0pt}\setlength{\partopsep}{0pt}\setlength{\parsep}{0pt}
  \item Fe, Co, and Ni are all magnetic at that temperature.
  \item Magnetic fluctuations are a well-known pair-breaking mechanism in conventional BCS superconductors,
        especially in A15-type materials that depend on strong electron--phonon coupling.
  \item The presence of up to 80\% magnetic elements makes it highly improbable for conventional
        superconductivity to survive.
\end{itemize}
\end{quote}

As stated above,
we define feasibility as the likelihood that,
given appropriate resources,
the claim can be reproduced
based on current scientific knowledge and current technology.
As such, it is an individual judgment;
there can therefore be variability over experts.
A good claim for the benchmark has a high level of agreement among experts.
Feasibility is represented using a 5-level Likert scale as described in \autoref{tab:feasibility}.
The explanations are expected to be brief and should describe the reasoning that led to the feasibility assessment as told to an expert.
Citations to literature and other evidence are included in the explanations.

\begin{table*}[t]
\caption{\textbf{Feasibility}. Definitions for the five levels of feasibility
that the annotators use to label scientific claims.}
\centering
\small
\begin{tabular}{p{0.05\textwidth} p{0.65\textwidth}}
\toprule
\textbf{Level} & \textbf{Description} \\
\midrule
+2 & Extremely likely to be feasible. Minor to no doubts. 95\% confident it’s feasible. \\
+1 & Somewhat likely to be feasible. Moderate doubts for but it might be possible. \\
~0 & Neither unlikely or likely to be feasible. No strong argument for or against. \\
-1 & Somewhat unlikely to be feasible. Moderate doubts against but cannot rule it out. \\
-2 & Extremely unlikely to be feasible. Significant doubts. 95\% confident it’s infeasible. \\
\bottomrule
\end{tabular}
\label{tab:feasibility}
\end{table*}

The \benchmarkname{} claims were restricted to four subdomains in materials science
with at least two SMEs recruited to develop and rate claims in each subdomain.
\autoref{tab:subdomains} lists the subdomains and the number of claims for each subdomain.
The subdomains were chosen based on 
the maturity of the subdomain
(which should support higher levels of agreement across experts)
and on SME availability.
\begin{table}
\caption{\textbf{Materials Science Subdomains}. Number of claims in each subdomain.}
\centering
\small
\begin{tabular}{p{0.74\columnwidth} p{0.11\columnwidth}}
\toprule
\textbf{Subdomains} & \textbf{Count} \\
\midrule
Lithium-Ion Batteries & 47 \\
Lightweight Aerospace Metal Alloys & 61 \\
A15 Superconductors & 29 \\
Thin-Film Semiconductors & 60 \\
\bottomrule
\end{tabular}
\label{tab:subdomains}
\end{table}

\begin{table*}[t]
\caption{\textbf{Dimensions.} The SMEs created claims that varied across these dimensions to represent different types of claims found in the wild.}
\centering
\small
\begin{tabular}{llp{4.1in}}
\toprule
Dimension & Type & Description\\
\midrule
Basic vs Applied &	Core & Degree to which claim rests on pure science (basic) vs. particular use case (applied)\\
Complexity &	Core &	Structural breadth of the claim, measured by number of subclaims\\
Modalities &	Reasoning &	Use of different evidence modalities, e.g., tables or figures\\
Tool Use &	Reasoning &	Use of tools as evidence, e.g., DB lookup, calculations, or simulations\\
\bottomrule
\end{tabular}
\label{tab:dimensions}
\end{table*}

Benchmark claims vary along four dimensions of interest,
as shown in \autoref{tab:dimensions}.
\textit{Core dimensions} are those that are intrinsic to the claim
and controlled for the benchmark during problem creation.
\textit{Reasoning dimensions} are those that are expected to vary in how feasibility assessment systems
arrive at their conclusions. For example, the SMEs used scientific computational tools when creating and evaluating some claims
so an automated feasibility system would likely benefit from using such a tool.

\subsection{Claim Creation}

Ten SMEs in the subdomains listed in Table~\ref{tab:subdomains} were recruited to write novel scientific claims for the benchmark.
A good benchmark claim is one that is difficult for current LLMs to assess
but that has high agreement among experts.
The SMEs had joint training sessions that introduced the purpose of the benchmark.
These sessions gave them an opportunity to write claims,
evaluate them using a baseline LLM system,
and then vary the content and difficulty of the claim to become more familiar with current LLM capabilities.

In early experiments, we learned that it was important to control the amount and types of ambiguity in the claims.
We wanted claims whose meaning would be understood and agreed upon by other materials experts.
Assumptions that were unlikely to be shared across experts needed to be explicitly included in the text of the claim.
For example, for claims that referred to low temperatures or high pressures,
we asked SMEs to provide precise values.

We also experimented with providing the experts with sample claims and associated feasibility scores created by an LLM that they could then modify.
We called these seed claims and tried multiple iterations with our SMEs.
By default, the LLM created claims that the SMEs found difficult to evaluate and actually slowed their claim creation rate.
To address this, we wrote technical biographies for each SME and restricted the LLM to producing claims that matched their area of expertise.
The SMEs found these seed claims much easier to understand and modify.
Overall, less than 10\% of claims in the collection were based on an LLM-generated seed claim.

We instructed the SMEs not to discuss their claims with each other
to preserve independence of judgment for our review process.
They were allowed to use any scientific or technical resource such as literature, databases, or simulations when writing claims.
However, we restricted their explanations to cite only open access literature, data, and tools.

After writing claims, each SME reviewed their own claims to fix issues with ambiguity.
Then a SME other than the author reviewed each claim,
provided a feasibility score,
and recorded questions and suggestions for the original claim author.
For the claims where there was disagreement in the sign
(e.g., the original SME rated the claim as -1 and the second SME rated it as +2),
the original claim author was asked to update the text of the claim
and to change its feasibility score based on feedback if a change was warranted.
These updated claims were shown again to the second SME to gather another feasibility annotation.

The quadratic weighted Cohen's kappa score after the first stage was 0.56;
after the second stage it improved to 0.74.

\subsection{Using the Benchmark}
The claims and explanations are represented in JSON and are available online.\footnote{\repo} 
Appendix~\ref{app:examples} includes examples of claims in the JSON format. We also provide a recommended JSON output format that our metrics code will read.
The metrics code automatically computes the accuracy of a system's feasibility scores.
As described in the next section, explanation grading requires SMEs.
We provide the rubric that we used for grading below.

\section{Metrics}
\label{sec:metrics}

A feasibility assessment is evaluated relative to the gold standard Likert score.
We describe suggested metrics in this section
and provide code implementing them at \repo.
Feasibility can be automatically scored while explanations require SMEs to grade their accuracy.

\subsection{Feasibility}

A feasibility assessment is evaluated with respect to its feasibility score and its explanation.
The primary metric for measuring alignment between 
the AI system-generated feasibility score and the human SME-specified gold standard is 
quadratic weighted Cohen's kappa.
While the Likert scale is discrete, the categories are not unrelated.
Thus we need a way to penalize differences that are far apart on the $-2$ to $2$ scale
more than those that are closer.
Weighted Cohen's kappa can produce counterintuitive
scores if the distribution of scores across categories is highly skewed.
For this reason, \benchmarkname{} claims have been selected so as to be roughly evenly distributed across feasibility levels.

\subsection{Explanations}
While we provide a gold standard explanation, there can be multiple equally correct explanations for a given claim.
In our experiments, we handled this by having SMEs manually grade automated system output.
We instructed SMEs to compare an explanation to the gold standard when appropriate.
If the two were not compatible,
the SME was to compare against an imagined idealized version of the explanation being graded.

Explanations are assessed along two axes: completeness and correctness.
We define a \textit{complete} explanation to be one that covers every main point that is needed to make the argument.
We define a \textit{correct} explanation to be one that is correct in each of its points,
includes appropriate citations,
and does not include irrelevant argument.
Inclusion of correct, properly cited explanation components
that are relevant but not part of an explanation deemed necessary by the SME
should not affect either score.

To measure correctness of the original evaluation,
SMEs selected one of the following categories for each explanation they assessed:
\begin{itemize}
\itemsep0em
    \item \textbf{Sound} -- All points are clear, accurate, and well-reasoned.
    \item \textbf{Mostly Sound} -- Generally accurate, with minor issues in clarity, detail, or logic.
    \item \textbf{Mixed} -- Contains some incorrect information or flawed reasoning.
    \item \textbf{Mostly Unsound} -- Significant errors in logic or science; difficult to follow.
    \item \textbf{Unsound} -- Major errors throughout; largely incoherent or deeply flawed.
\end{itemize}

To measure completeness of the original evaluation,
SMEs selected one of the following categories for each explanation they assessed:
\begin{itemize}
\itemsep0em
    \item \textbf{Complete} -- Includes all key points and valid supporting citations.
    \item \textbf{Strong} -- Misses a minor point or lacks one citation.
    \item \textbf{Adequate} -- Missing at least one key point; citations are incomplete.
    \item \textbf{Weak} -- Omits several key points; most citations are missing or weak.
    \item \textbf{Incomplete} -- Major omissions; does not resemble the gold standard; citations are missing or invalid.
\end{itemize}

An explanation is correct to the extent that it accurately conveys information
useful for understanding one or more aspects of the explanation.
Components that are correct are useful to the extent that they help with understanding
whether a claim is feasible.

An explanation is deemed complete if it covers all points the SME believes to be essential
for a good explanation.
This judgment will be different for explanations that have negative and non-negative Likert scores.
An explanation of an infeasible claim is complete if it demonstrates one reason the claim is infeasible;
it is not necessary to include every reason the claim is infeasible.
In contrast, an explanation of a claim with non-negative Likert score
must include enough information to validate all aspects of feasibility deemed to be crucial
by the SME.
Which aspects are crucial is of course an opinion,
which is one reason that feasibility is considered to be an opinion, not a fact.

In making their judgments,
SMEs were encouraged to consider at least three aspects of feasibility:
theoretical limits, technological limits, and empirical evidence.
Claims that are likely to violate theoretical or technological limits
will usually receive a negative Likert feasibility score.
Claims that conflict with most empirical evidence will usually receive a negative Likert feasibility score,
while those that are supported by most empirical evidence will usually receive a positive score.

Once the usefulness and comprehensiveness of each problem has been assessed,
we can calculate correctness and completeness scores for an evaluation run
by assigning points to each response category,
summing across all problems,
and normalizing by the score that would have been achieved
had every question been answered with the highest possible category.
For example, if we assign a value of five to the top category
down to one for the lowest category,
then these scores can be calculated as
\[
\mathrm{C}(R) = \frac{\sum_{i=1}^{k}(S(R_i) - 1)}{4k}
\]
where $k$ is the number of claims in the test set,
$R_i$ is the category assigned to the $k$th claim by the system,
$S$ is the numeric value assigned to the category,
and C is either ``correctness'' or ``completeness.''
Given correctness and completeness scores,
we can compute a single numeric score as the harmonic mean (often called $F_1$) of the two scores:
\[
F_1 = \frac{2 \times \mathrm{correctness} \times \mathrm{completeness}}{\mathrm{correctness}+\mathrm{completeness}}
\]

\section{Baseline Performance}

\begin{figure}
    \centering
    \includegraphics[width=\columnwidth]{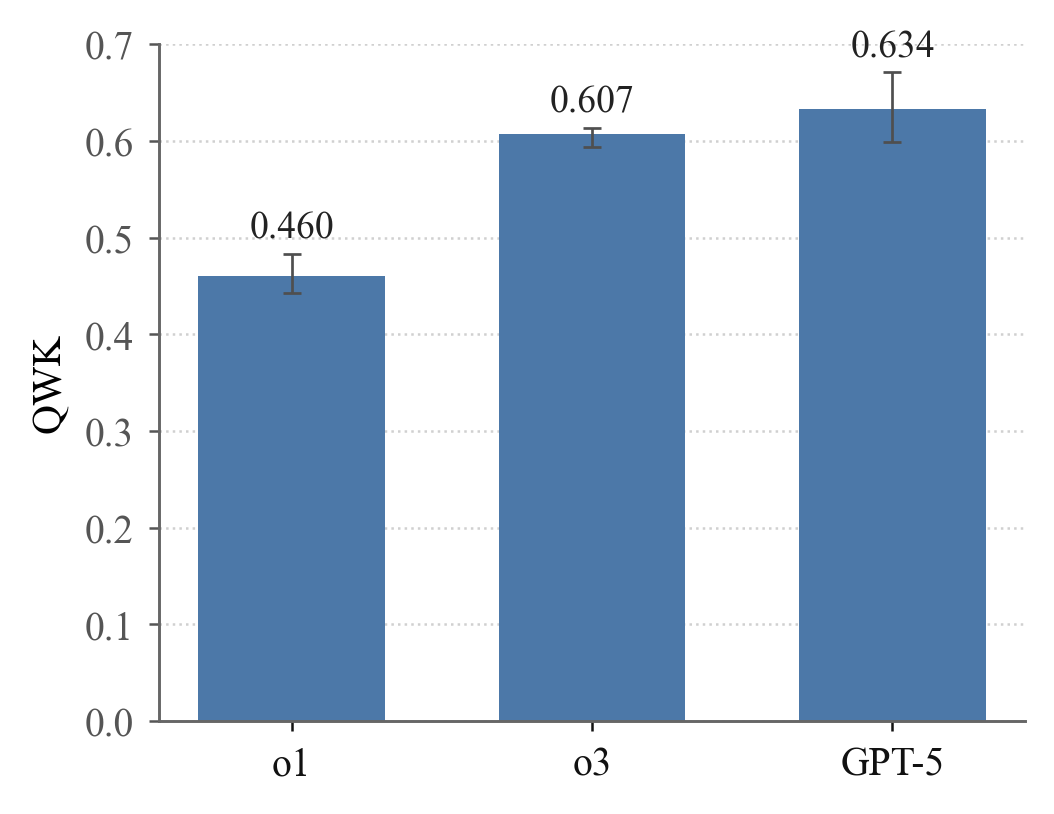}
    \caption{Quadratic weighted kappa for baseline models evaluated on SFBench. 
    Error bars show the maximum and minimum across three runs. All models have reasoning effort set to high. Performance increases with newer model bases. A kappa of zero or below reflects no agreement. 
    }
    \label{fig:baseline-results}
\end{figure}

We evaluate the performance of a standard LLM system on this benchmark as our baseline. We use a commercial LLM base with a single prompt including manually-curated instructions and few-shot examples\footnote{The full prompt, provided to all model bases, is available in \autoref{app:baseline-prompt}; we manually explored wording variations to arrive at a well-performing prompt. The incorporated few-shot examples were created by the materials science SMEs in preliminary problem generation sessions.} and receive a single response from the model providing the feasibility score and explanation together. This system relies only on the models' parametric knowledge with no web search capabilities. The base LLM is interchangeable; in our experiments, we evaluate o1, o3, and GPT-5 as representative samples of the state of the art over time.

The effectiveness of three LLM baselines can be seen in \autoref{fig:baseline-results}, where we report the mean performance over three runs of each system. Baseline effectiveness improves as the overall power of the LLM increases. 

We analyzed the stability of these model bases by calculating Krippendorff's $\alpha$ for the three runs of each LLM: o1 achieved $\alpha = 0.91$, o3 achieved 0.81, and GPT-5 achieved 0.90, all of which are considered reliable.

\section{Discussion}

As we developed the benchmark and evaluated automated feasibility systems,
we identified challenges in evaluating AI systems
that approach human levels of performance that deserve further discussion.

\paragraph{Difficulty of creating novel feasible claims.}
Our SMEs relied on the mental models they developed over years of experience
to extrapolate from current claims about known materials under common conditions
to create novel claims.
They found it significantly easier to create challenging infeasible claims than feasible claims.
For example, for an infeasible claim they could increase the percentage of an element
past the point of structural changes that would affect a material property such as brittleness.
Then they could devise a scenario requiring a ductile material
and claim that the novel material, which is actually brittle, was used.
There would usually be no scientific literature about this novel material,
requiring the automated feasibility assessment system to reason about how that particular composition would affect the material's required properties.
This type of approach achieved strong agreement on the feasibility score among annotators.

In contrast, it is more difficult to begin with a known feasible claim
and modify it so that it is at once novel,
difficult for AIs to assess,
and has agreement among SMEs.
Novel feasible claims in the subdomains we used tend to be obviously feasible
or require experimental evidence for SMEs to agree on their feasibility.

\paragraph{Specialization of human experts.}
To construct the benchmark, we recruited subject-matter experts (SMEs)
from four materials science sub-domains,
with at least two experts representing each sub-domain.
We expected specialization would be a challenge
as a materials scientist focused on superconductors, for example,
is unlikely to reliably assess claims about alloys.
This problem held even within sub-domains.
SMEs tended to generate claims closely aligned with their own narrow expertise,
which reduced the likelihood that another SME from the same sub-domain
would have sufficient specialization to evaluate those claims confidently.
This observation led us to ask SMEs to write claims that they believed
another expert in their sub-domain could reasonably assess.
This phenomenon highlights a contrast between human and AI systems:
human experts often achieve very high accuracy within narrowly defined topics
but only moderate accuracy outside their specialization,
whereas AI systems have the potential to maintain a higher
baseline level of competence across a broader range of topics.

\paragraph{Importance of retrieval versus reasoning.}
Some of the generated claims were inspired by recent papers that the SME had read.
They took a claim from the paper and modified it to make it more or less feasible.
For such claims, retrieving that exact paper is likely advantageous
for assessing that claim.
More generally, it is an open question of the degree to which
a correct feasibility assessment depends on retrieving particular evidence
versus pure reasoning ability.
Further experiments could be run with expert-identified literature to compare system effectiveness with and without these documents.

\section{Conclusion}
We introduced \benchmarkname{}, a benchmark for evaluating scientific feasibility assessment systems. The dataset consists of expert-authored, de novo scientific claims paired with feasibility judgments and open-ended explanations.
By avoiding extracted claims and multiple-choice formats, \benchmarkname{} emphasizes reasoning over memorization, providing a unique benchmark for evaluating automated scientific judgment capabilities.
\benchmarkname{} supports studying scientific reasoning, explanation generation, and the balance between retrieval and inference in AI systems.

Baseline evaluations with LLMs show that performance improves with model capability but remains below expert agreement levels. This makes it an interesting dataset for exploring interpretation of claims, multi-step reasoning and evidence weighing.
Future work includes expanding to additional scientific areas, improving automated evaluation of explanations, and increasing use of automated experimentation.

\section*{Acknowledgments}

We thank the subject matter experts who contributed to the \benchmarkname{} dataset, including Adrian Podpirka, Andy Lennon, Christopher Stiles, James Hays-Wehle, Jarod Gagnon, Karun Kumar Rao, Kyle McElroy, Lisa Pogue, Nam Le, Rachel Altmaier, and Tanner Hamann.
Their expertise was essential in developing the scientific claims, providing feasibility assessments, and grading  explanations.  

\bibliography{hugo}

\appendix
\newpage
\onecolumn

\section{Baseline Prompt}\label{app:baseline-prompt}

This appendix provides the full prompt used in our baseline experiments, including task instructions and few-shot examples.
We include this to illustrate the level of guidance provided to the model.
In the prompt, \texttt{\{\{claim\}\}} represents a single input claim.

\lstset{
  basicstyle=\ttfamily\small,
  breaklines=true,
  breakatwhitespace=true,
  columns=fullflexible,
  keepspaces=true
}

\begin{lstlisting}[language={}]
Your overall goal is to evaluate scientific claims for feasibility. 
Feasibility is defined as the likelihood that the claim could be reproduced by experts based on current scientific knowledge and current technology, given appropriate resources. It is an opinion of an expert. By "reproduced", we mean that a skilled team with plenty of resources using the latest science and technology would be able to accomplish the claim.
When information is missing or ambiguous, consider plausible alternative values and interpretations (e.g., different pressures, temperatures, scales) if they exist, and justify any assumptions you make. Combine the feasibility scores for these alternative interpretations into a single final score, based on the likelihood of each interpretation.

You will be shown a single scientific claim and asked to assess its feasibility.

Score the feasibility of the claim on an integer scale between -2 and 2. Use the following scale:
+2: Extremely likely to be feasible. Minor to no doubts. 95% confident it's feasible.
+1: Somewhat likely to be feasible. Moderate doubts for but it might be possible.
0: Neither unlikely or likely to be feasible. No strong argument for or against.
-1: Somewhat unlikely to be feasible. Moderate doubts against but cannot rule it out.
-2: Extremely unlikely to be feasible. Significant doubts. 95% confident it's infeasible.

At the end, output exactly these sections, each on a new line with exactly the label shown:

Feasibility: integer feasibility score in [-2..2] inclusive
Explanation: a bulleted or numbered list of logical inferences explaining and deriving the feasibility of the claim including cited hyperlinks

Do not add other sections beneath these headers.

Some examples of this task are as follows:
INPUT:
Claim: La$_{4}$H$_{23}$ has been found to exhibit superconductivity at 120 GPa.

OUTPUT:
Feasibility: 2
Explanation: 1. La$_{4}$H$_{23}$ possesses a hydrogen-rich lattice capable of strong electron-phonon interactions. (https://arxiv.org/pdf/2307.13067)
2. Ab initio simulations predict metallic behavior and favorable coupling constants. (https://arxiv.org/pdf/2307.13067, https://pubs.rsc.org/en/content/articlehtml/2025/tc/d5tc02600h)
3. Known constraints (thermodynamic stability, vibrational modes, synthesis conditions) are satisfied under high pressure. (https://arxiv.org/pdf/2307.13067, https://arxiv.org/pdf/2308.02977)
4. There are no known physical blockers at that pressure-temperature range. (https://arxiv.org/pdf/2307.13067)
5. Related materials in the lanthanum-hydride family have shown similar behavior. (https://arxiv.org/pdf/2308.02977, https://pubs.rsc.org/en/content/articlehtml/2025/tc/d5tc02600h, https://pubmed.ncbi.nlm.nih.gov/31118520/)
6. Based on the above evidence, La$_{4}$H$_{23}$ is scientifically feasible as a superconductor at 120 GPa.


INPUT:
Claim: Gallium dopant enhances fracture toughness of iron alloy.

OUTPUT:
Feasibility: -2
Explanation: 1. Gallium melts at 29.8$^{\circ}$C, so it will be a liquid at fabrication temperatures. (https://tsapps.nist.gov/srmext/certificates/archives/8174.pdf)
2. Liquids flow into microscopic crevices, especially along grain boundaries. (https://www.sciencedirect.com/science/article/pii/S1359645404005543)
3. These boundaries become thin films of a Ga-rich phase. (https://www.sciencedirect.com/science/article/abs/pii/S1359645408008495)
4. This causes liquid-metal embrittlement which makes the alloy more susceptible to fracture. (https://link.springer.com/article/10.1007/s11661-021-06256-y, https://www.osti.gov/pages/biblio/1479997)
5. Increased susceptibility to fracture is in direct disagreement with the enhanced fracture toughness claim. Therefore, this claim is infeasible.

Now generate the output for:
INPUT:
Claim: {{claim}}.

OUTPUT:
\end{lstlisting}
\newpage
\onecolumn

\section{Example Claims}\label{app:examples}

\benchmarkname{} includes \numclaims{} claims in materials science.  Each claim includes:
\begin{itemize}
    \item The text of the claim (a sentence or short paragraph)
    \item The gold standard feasibility score (a Likert scale from -2 to +2)
    \item An example explanation supporting that feasibility score (several statements in text with supporting cites)
    \item A tag assigning the claim to one of four subdomains (alloys, batteries, semiconductors, superconductors)
    \item A set of tags assigning dimension categories to the claim (applied, tool used, modalities, complex)
    \item Some typical metadata such as problem ID (a unique number for the claim) and format version number
\end{itemize}

\noindent Below are two example claims in the JSON format.

\begin{lstlisting}[language={}]
{
 "type": "gold standard", "format_version": "1.0", "problem_id": "problem_652", "problem_version": "1.0", "domain": "materials", "subdomain": "alloys", 
 "claim": "An aluminum alloy with thermally stable nanometer-scale dispersoids can maintain microstructures with grain sizes of 10 microns or less during superplastic forming, enabling elongation beyond 300% without premature loss of ductility.", 
 "likert_score": 2, 
 "explanation": [{"text": "10 microns is already achievable with commercial alloys, and there is no reason to believe that the addition of nano-dispersoids would increase that grain size.", "evidence": ["evidence_0_0", "evidence_0_1"]}], 
 "evidence": {"evidence_0_0": {"type": "url", "source": "https://apps.dtic.mil/sti/tr/pdf/ADA201941.pdf"}, "evidence_0_1": {"type": "url", "source": "https://apps.dtic.mil/sti/tr/pdf/ADA209678.pdf"}}, 
 "tags": {"applied": true, "tool_used": null, "modalities": false, "complex": false}
}
 
{
 "type": "gold standard", "format_version": "1.0", "problem_id": "problem_653", "problem_version": "1.0", "domain": "materials", "subdomain": "semiconductors", 
 "claim": "CZTSSe thin films can be reliably grown in a sufficiently phase-pure form over large-area flexible substrates in a roll-to-roll process using sputter deposition under tuned sulfur/selenium chemical potential conditions. This is done in a hybrid cosputtering setup using 3 targets and followed by a subsequent sulfurization and/or selenization heat treatment to improve cell efficiencies.", "artifacts": [], 
 "likert_score": 1, 
 "explanation": [{"text": "This is a relatively common approach to getting CZTS but the substrate flexibility gives a little development nuance. The trick is that one rarely gets the right amount of control over the sulfur and selenium profile without a subsequent heat treatment. CIGS cells are regularly deposited on flexible substrates in a roll to roll fashion (on the same back contact with similar engineering of the back interface required and Na content) so one would not expect significant development to be required to do this.", "evidence": ["evidence_0_0"]}], 
 "evidence": {"evidence_0_0": {"type": "SME Knowledge", "source": "SME knowledge"}}, 
 "tags": {"applied": false, "tool_used": null, "modalities": false, "complex": true}
}
\end{lstlisting}
\end{document}